\documentclass{article} 
\usepackage{iclr2021_conference,times}

\usepackage{hyperref}
\usepackage{url}
\usepackage{booktabs}       
\usepackage{amsfonts}       
\usepackage{nicefrac}       
\usepackage{microtype}      
\usepackage{graphicx} 
\usepackage{amsmath,soul}
\usepackage{amssymb}
\usepackage[font=small,skip=0pt]{subcaption} 
\usepackage{balance}
\usepackage{wrapfig}
\usepackage[ruled,linesnumbered,vlined]{algorithm2e}
\usepackage[normalem]{ulem}
\usepackage{bm}
\usepackage{bbm}
\usepackage{mathtools}
\usepackage{enumitem}
\usepackage{thmtools}
\usepackage{thm-restate}
\usepackage{cleveref}
\renewcommand\cite{\citep}
\DeclareMathOperator{\E}{\mathbb{E}} 
\DeclareMathOperator{\R}{\mathbb{R}} 

\title{Dependency Structure Misspecification in Multi-Source Weak Supervision Models}

\author{%
  Salva Rühling Cachay \\
   Technical University of Darmstadt\\
  \texttt{salvaruehling@gmail.com} \\
  \And
  Benedikt Boecking \\
  Carnegie Mellon University \\
  \And
  Artur Dubrawski \\
  Carnegie Mellon University 
}

\iclrfinalcopy 
\begin{document}

\maketitle
\vspace{-6mm}
\begin{abstract}
    Data programming (DP) has proven to be an attractive alternative to costly hand-labeling of data.
    In DP, users encode domain knowledge into \emph{labeling functions} (LF), heuristics that label a subset of the data noisily and may have complex dependencies. A label model is then fit to the LFs to produce an estimate of the unknown class label.
    The effects of label model misspecification on test set performance of a downstream classifier are understudied. This presents a serious awareness gap to practitioners, in particular since the dependency structure among LFs is frequently ignored in field applications of DP.
    We analyse modeling errors due to structure over-specification.
    We derive novel theoretical bounds on the modeling error and empirically show that this error can be substantial, even when modeling a seemingly sensible structure. 
\end{abstract}

\vspace{-3mm}
\section{Introduction}\label{sec:intro}
\vspace{-2mm}
Annotating large datasets for machine learning is expensive, time consuming, and a bottleneck for many practical applications of artificial intelligence. Recently, data programming, a paradigm that makes use of multiple weak supervision sources, has emerged as a promising alternative to manual data annotation~\cite{DP}. In this framework, users encode domain knowledge into weak supervision sources, such as domain heuristics or knowledge bases, that each noisily label a subset of the data. A generative model over the sources and the latent true label is then learned. One can use the learned model to estimate \emph{probabilistic} labels to train a \emph{downstream model}, replacing the need to obtain ground truth labels by manual annotation of individual samples.\\
In practice, the sources of weak labels that users define often exhibit statistical dependencies amongst each other, e.g. sources operating on the same or similar input (some examples can be found in Table \ref{imdb_example_deps}).
Defining the correct dependency structure is difficult, thus a common approach in popular libraries~\cite{Drybell, Snorkel} and related research~\cite{SkeneModel, anandkumar2014tensor, Automatic-Lf-gen, IWS} is to ignore it.
However, the implications of this assumption on downstream performance have not been researched in detail. Therefore, in this paper we take steps towards gaining a better understanding of the trade-offs involved.

\textbf{Contributions} $\;$ We present novel bounds on the label model posterior and the downstream generalization risk that are explicitly influenced by misspecified higher-order dependencies. We also introduce three new higher-order dependency types, which we name \emph{bolstering}, \emph{negated}, and \emph{priority} dependencies. Lastly, we empirically show that downstream test performance is highly sensitive to the user-specified dependencies, even when they make sense semantically. The finding suggests that in practice, it is advisable to only carefully model a few, if any, dependencies.

\vspace{-2mm}
\section{Related Work}
\vspace{-2mm}
\textbf{Data programming} $\;$ While the original data programming framework \cite{DP} is based on a factor graph that support the modeling of arbitrary dependencies between LFs, more recent methods for solving for the parameters of the label model only support the modeling of pairwise correlations \cite{Snorkel, Multitask, triplets} --- as such, losing some of the expressive power of data programming. The former extends data programming to the multi-task setting by exploiting the graph structure of the inverse covariance matrix among the sources \cite{Multitask} --- in particular the fact that an entry is zero when there is no edge between the corresponding sources in the graphical model \cite{structureLearning2Theory}. 
The latter finds a closed-form solution for a class of binary Ising models by using triplet methods \cite{triplets}.
Our experimental findings suggest that practitioners may indeed benefit from simply ignoring higher-order dependencies.\\
\textbf{Structure learning} $\;$ In order to automatically learn the structure between these sources, previous work optimizes the marginal pseudolikelihood of the noisy labels \cite{structureLearning1}, or makes use of robust PCA to denoise the inverse covariance matrix of the sources labels into a graph structured term \cite{structureLearning2}. A different approach, infers the structure through static analysis of the weak supervision sources code definitions and thereby reduces the sample complexity for learning the structure \cite{SL-static-analysis}. In our experiments, we show that such methods should be carefully used, and may in fact lead to downstream performance losses.\\
\textbf{Model misspecification} $\;$On the side of work on model misspecification, \cite{misspecified_MLE} establishes that the Maximum Likelihood Estimator of a misspecified model is a consistent estimator of the learnable parameter that minimizes the KL divergence to the true distribution -- if that optimal, misspecified parameter is globally identifiable.
In an interesting result, \cite{misspecified_Gaussian_PGM} show that the KL divergence between a multivariate Gaussian distribution and a misspecified (by at least a single edge) Gaussian graphical model is bounded by a constant from below. It emphasizes the need to correctly select the model's edge structure, since otherwise the fitted distribution will differ from the true one in terms of KL divergence.

\vspace{-2mm}
\section{Problem setup}
\vspace{-2mm}
For this work we use the data programming framework as introduced in~\cite{DP}, where the premise is that experts can model any higher-order dependency between labeling functions. We extend it by \emph{negated}, \emph{bolstering}, \emph{priority} dependencies (definitions in the appendix \ref{sec:factor_defs}), e.g. the latter encoding the notion that one source's vote should be prioritized over the one from a noisier source. The additional dependency types are motivated by our selected LF sets that contain dependencies, such as the ones in Table \ref{imdb_example_deps}, that we could not express before. Newer weak supervision models and model fitting approaches often only allow for pairwise correlation dependencies to be modeled  \cite{Multitask,triplets}.\\
Let $(x,y) \sim \mathcal{D}$ be the true data generating distribution and for simplicity assume that $y \in \{-1, 1\}$. As in~\cite{DP}, users provide $m$ labeling functions (LFs) $\lambda = \lambda(x) \in \{-1, 0, 1 \}^m$, where $0$ means that the LF abstained from labeling. Following~\cite{DP}, we model the joint distribution of $y,\lambda$ as a factor graph. 
To study model misspecification we compare two label models, $p_{\theta}$ for the conditional independent case  and $p_\mu$ which models $M$ higher-order dependencies: 
\begin{align}
    p_\mu(\lambda, y) 
    = \frac{1}{Z_\mu} \exp \left(\mu^T \phi(\lambda, y) \right)
    &=  Z_{{\mu}}^{-1} \exp \left(\mu_1^T \phi_1(\lambda, y) + \mu_2^T \phi_2(\lambda, y)  \right) 
    , \qquad \mu \in \R^{m + M} \\
    p_{\theta}(\lambda, y) &= Z_{{\theta}}^{-1} \exp \left(\theta^T \phi_1(\lambda, y) \right), \qquad \theta \in \R^{m},
\end{align}
where $\phi_1(\lambda, y) = \lambda y$ are the accuracy factors, $\phi_2(\cdot)$ are arbitrary, higher-order dependencies and $ Z_{{\theta}}^{-1},  Z_{{\mu}}^{-1}$ are normalization constants. We assume w.l.o.g. that factors are bounded $\le1$.  
Note that for ease of exposition, the models above do not model the labeling propensity factor (also known as LF coverage) . Since this factor does not depend on the label, the corresponding terms would cancel out in the quantities studied below (same goes for dependencies that do not depend on $y$, e.g. the \emph{similar} factor from \cite{DP}).
\vspace{-2mm}
\section{Theoretical analysis}
\vspace{-2mm}
\paragraph{Bound on the label model posterior under model misspecification} 
We now state our bound on the probabilistic label difference (which we prove in the appendix \ref{sec:proofs}):
\begin{equation} \label{BoundProbs}
    \left| p_\mu(y \;|\lambda) - p_\theta(y \;|\lambda) \right|
    \le
    \frac{1}{2} ||\mu_1 - \theta||_1 + \frac{1}{4} ||\mu_2||_1
\end{equation}
This is an important quantity of interest since the probabilistic labels are used to train a downstream model. Unsurprisingly then, this quantity reappears as a main factor controlling the KL divergence and generalization risk, see below. \\
Suppose that the downstream model $f_w : \mathcal{X} \rightarrow \mathcal{Y}$ is parameterized by $w$, and
that to learn $w$ we minimize a, w.l.o.g., bounded noise-aware loss function $L(f_w(x), y) \in [0, 1]$ (that acts on the probabilistic labels).
If we had access to the true labels, we would normally try to find the $w$ that minimizes the risk: 
$w^* = \arg \min_w R(w) =  \arg \min_w  \E_{(x,y) \sim D} \left[ L ( f_w(x), y)\right]$.
Since this is not the case, we instead will get the parameter $\hat w$ that minimizes the (empirical) noise-aware loss.
\vspace{-2mm}
\paragraph{Bound on the generalization risk under model misspecification}
While \cite{triplets} provide a bound on the generalization error that accounts for model misspecification -- the label model being a less expressive Ising model (i.e. no higher-order dependencies may be modeled) -- the part of the bound corresponding to the misspecification is a somewhat loose KL-divergence term between the true and the misspecified models. If we instead assume that there exists a label model for some optimal parameterization of factors (not necessarily the ones that are actually modeled) that is equivalent to the true data generating distribution, we can provide a more meaningful and tight bound. \\
As in \cite{DP, Multitask}, assume that 1) there exists an optimal parameter of either $p_\mu$ or $p_\theta$ (say, $\mu^*$ with label model $p_{\mu^*}$) such that sampling $(\lambda, y)$ from this optimal label model is equivalent to $(\lambda, y) \sim \mathcal{D}$; and 2) the label $y$ is independent of the features used for training $f_w$ given the labeling function outputs $\lambda$.
Differently than \cite{DP, Multitask}, the factors that are actually modeled may differ from the ones of the optimal label model.
By adapting the proof of theorem 1 in~\cite{Multitask} to our case where we incorrectly use a misspecified model (say, $p_\theta$), we can bound the generalization risk as follows
\begin{equation} \label{BoundRisk}
    R(\hat w) - R(w^*) 
    \le 
    \gamma 
    + 2||\mu_1^* - \theta||_1 + ||\mu_2^*||_1,
\end{equation}
where $\gamma$ is a bound on the empirical risk minimization error
, which is not specific to our setting.
We can bound the KL divergence of the two models by a similar quantity, see the appendix \ref{KL-div_bound}.

\paragraph{Interpretation} 
These bounds naturally involve the accuracy parameter estimation error $||\mu_1 - \theta||_1$ and the learned strength $s := ||\mu_2||_1$ of the dependencies only modeled in $p_\mu$. 
Note that if we assume that the model with dependencies $p_\mu$ is the true one we can interpret $s$ as the magnitude of the dependencies that the independent model $p_\theta$ failed to model. If on the other hand we associate the misspecified model with $p_\mu$, then we can interpret this quantity $s$ as the (dependency) parameter excess that was incorrectly learned by $p_\mu$.
The presented bounds are tighter than the ones from~\cite{Multitask} for the L1 norm, while in addition accounting for model misspecification.
\section{Experiments}
\begin{figure}
\begin{subfigure}{.5\textwidth}
  \centering
  \includegraphics[width=.85\linewidth]{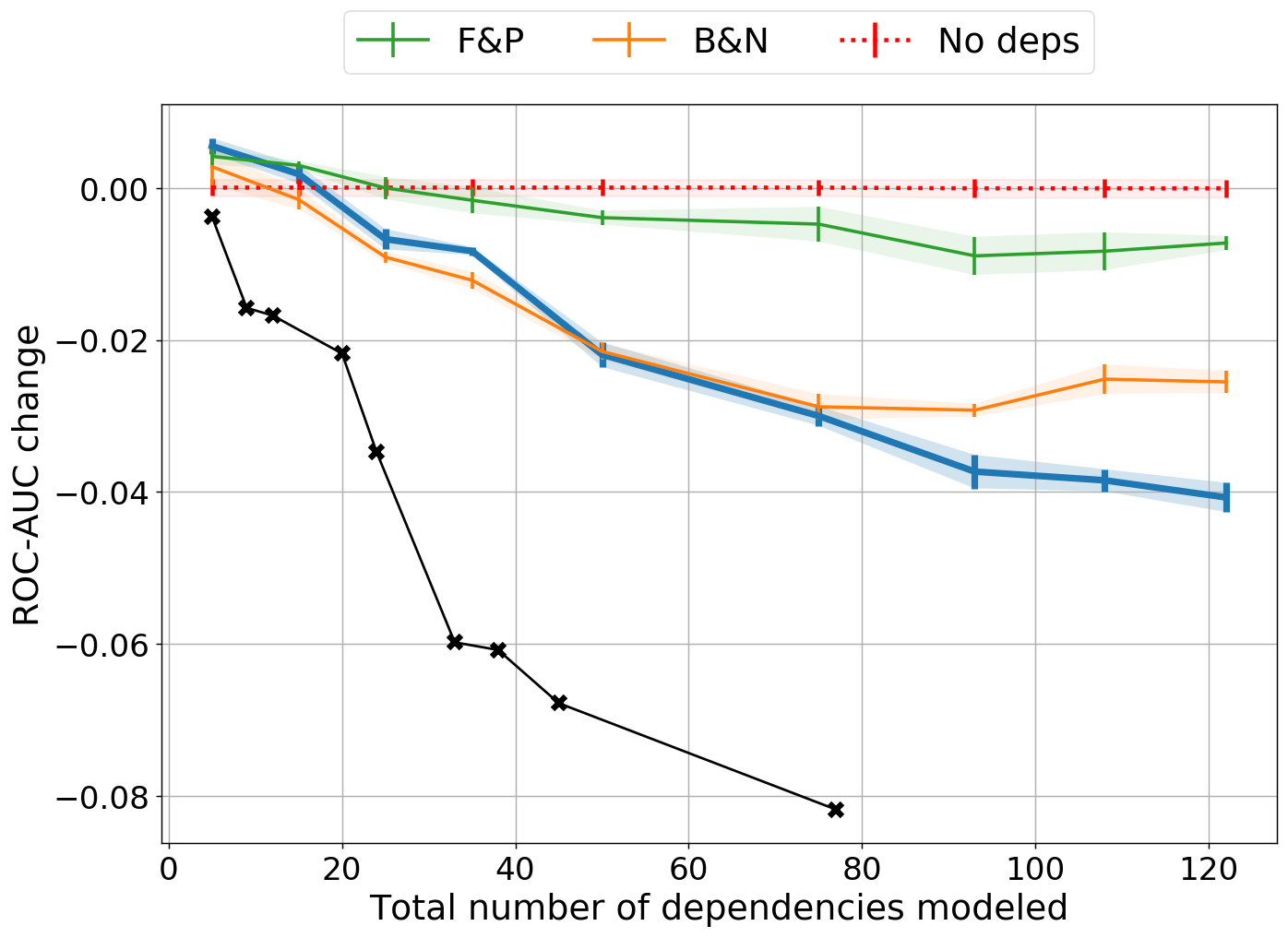}
  \caption{IMDB ($0.81$)}
  \label{fig:imdb}
\end{subfigure}%
\begin{subfigure}{.5\textwidth}
  \centering
  \includegraphics[width=.85\linewidth]{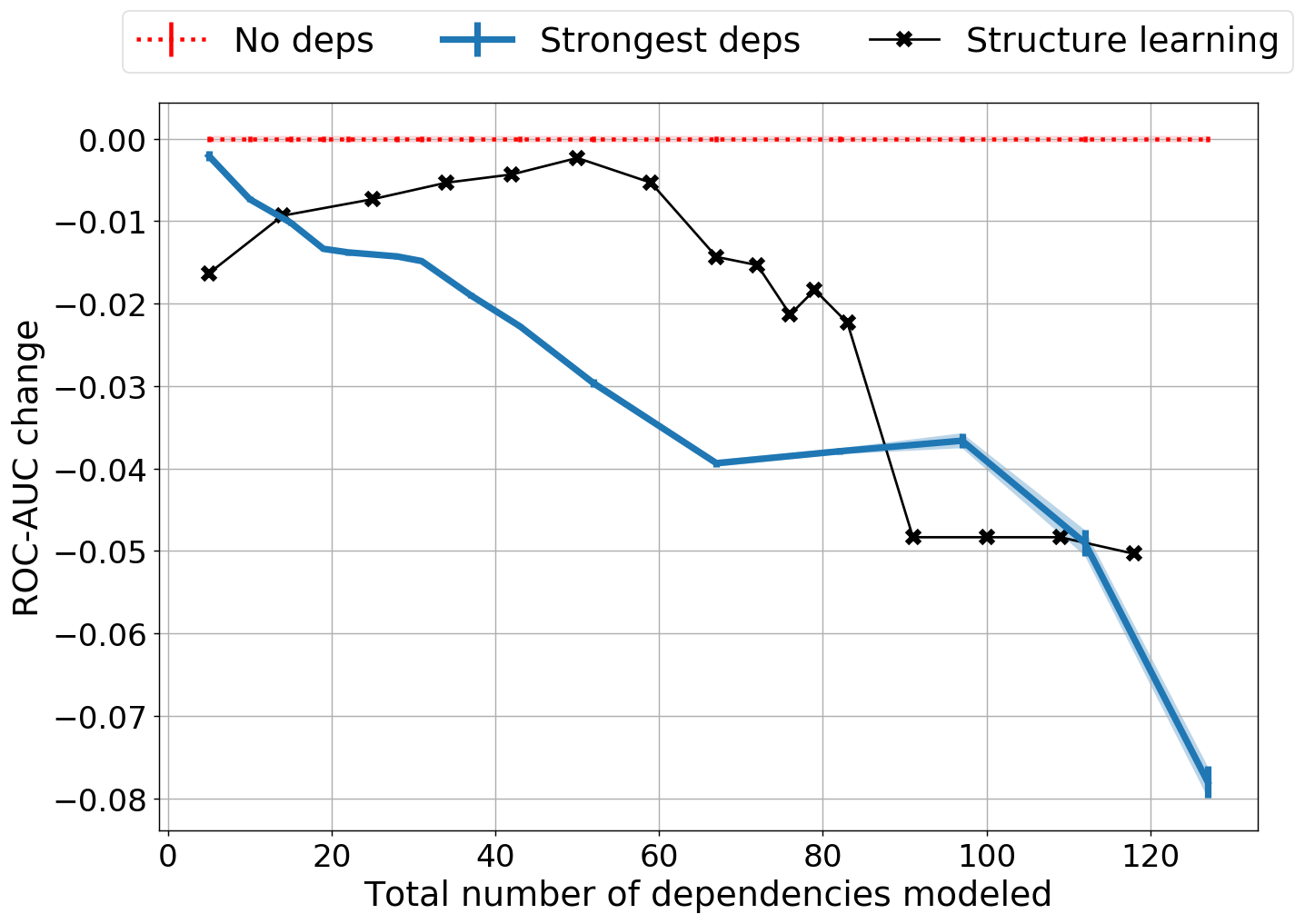}
  \caption{Bias Bios: professor or teacher ($0.91$)}
  \label{fig:biasbios}
\end{subfigure}
  \caption{Modeling more than a handful dependencies (as the ones in Table~\ref{imdb_example_deps}) significantly deteriorates ROC-AUC downstream performance as compared to simply ignoring them by up to 4 and 8 points. This effect intensifies as we model more dependencies. In brackets, the baseline score for the independent model (`No deps`).
  \texttt{B\&N} means that we only model the \emph{bolstering} and \emph{negated} dependencies, while 
  \texttt{F\&P} means that we only model the \emph{fixing} and \emph{priority} dependencies. \texttt{Structure learning} means that we model the \emph{similar}, \emph{fixing} and \emph{reinforcing} dependencies returned by the method from \cite{structureLearning1} for different threshold hyperparameters (the lower, the more dependencies are modeled).
  }
  \label{fig:fig_all_lfs}
\end{figure}
\vspace{-3mm}
\subsection{Proxy for finding true dependencies in real datasets}
\vspace{-2mm}
\label{real_datasets_method}
The underlying \emph{true} structure between two real labeling functions $\lambda_j,\lambda_k$ is, of course, unknown.
However, by using true training labels (solely for this purpose) together with the observed LF votes, we can compute resulting factor values for each data point $i$, to observe empirical strength of dependency factor $l$  over a training set: $v^l_{j,k}=\sum_{i}\phi^{l}(\lambda(x_i)_j,\lambda(x_i)_k,y_i)$.
Sorting dependencies $l$  according to $v^l_{j,k}$ in descending order, we then choose to model the top $d$ dependencies. These are the dependencies for which the true labels provide the most evidence of being correct. 

\subsection{Performance deterioration due to structure over-specification} 

For the following experiment we use the IMDB Movie Review Sentiment dataset 
consisting of $n=25$k training and test samples each~\cite{IMDB} and manually select a set of $m=135$ sensible LFs that label on the presence of a single word or a pair of words (i.e. uni-/bi-gram LFs).
In addition we use the Bias in Bios dataset \cite{biasBios}, and focus on the binary classification task introduced in \cite{IWS} that aims at distinguishing the biographies of professors and teachers ($n = 12 294, m=85$). We deliberately choose unigram and bigram LFs so as to create dependencies we expect to help with downstream model performance, e.g. by adding negations (e.g. "not worth") of unigrams (e.g. "worth") that we expect to fix the less precise votes of the latter when both do not abstain.\\
\begin{table}
\caption{The strongest and weakest dependencies among the IMDB LFs, as measured by their factor value $v^l_{j,k}$ computed with respect to the true labels. In the experiment from Fig.\ref{fig:fig_all_lfs}, we repeatedly add weaker dependencies to the set of dependencies used for learning the label model, starting with the strongest ones below.}
    \label{imdb_example_deps}
    \vspace{-2mm}
    \centering
    \begin{tabular}{@{} *4l @{}}
    \toprule
    LF$_j$     & LF$_k$ & factor type $l$ & factor value $v^l_{j,k}$\\ 
    \midrule
    best & great & bolstering & 801  \\
    bad & don't waste & bolstering & 110  \\
    original & bad & priority & 327 \\
    recommend & terrible & priority & 53 \\
    worth & not worth & fixing & 238  \\
    great & nothing great & fixing & 15 \\
    worth & not worth & negated & 219  \\
    special & not special & negated & 8\\
    recommend & highly recommend & reinforcing & 226   \\ 
    bad &  absolutely horrible & reinforcing & 7 \\
    \bottomrule
    \end{tabular}
\end{table}
We choose different $d \in \{1, 3, 5, \dots, 40\}$ and then model the strongest $\le d$ dependencies of each factor $l$ according to $v^l$. 
An example of the strongest and weakest dependencies for the IMDB dataset is shown in Table~\ref{imdb_example_deps}. For the Bias in Bios experiment, an example of a strong bolstering dependency is that the term `PhD' appears in addition to the term `university'. 
We report the test set performance of a simple 3-layer neural network trained on the probabilistic labels, averaged out over 100 runs.
While for IMDB we observe a marginal boost ($< 0.005$) in performance when modeling the strongest $d=1,3$ dependencies of each factor ($5, 15$ in total), the main take-away is the following:\\
We find that modeling more than a handful of dependencies significantly \emph{deteriorates} the downstream end classifier performance (by up to 8 ROC-AUC points) as compared to simply ignoring them (Fig.~\ref{fig:fig_all_lfs}).
The performance worsens as we increase $d$, i.e. as we model more, slightly weaker, dependencies. We reiterate that these additional dependencies still, semantically, make sense (as depicted in Table \ref{imdb_example_deps}, where the weakest ones are modeled only for the case where the total number of dependencies $=122$). Using the structure learning method from \cite{structureLearning1} to infer the dependency structure results in worse test performances too.

\section{Discussion and Future Work}
\vspace{-1.5mm}
\textbf{Discussion} $\;$ Even though this result and insight is highly relevant for practitioners, it has, to the best of our knowledge, not been explored in detail. It may come as a surprise that modeling seemingly sensible dependencies can significantly deteriorate the targeted downstream model performance.
We hypothesize that this is due to the true model being close to the conditionally independent case and in our presented bounds, we indeed see that they become looser as more incorrect dependencies are modeled. In addition, more complex models often suffer of a higher sample complexity, as is also briefly noted in ~\cite{triplets}.
This suggests that practitioners may 1) indeed be best served, at first, by simply ignoring (higher-order) dependencies; and 2) need to be careful when specifying dependencies, either by hand or through structure learning algorithms, which emphasizes the need for a small ground-truth labeled validation set to compare the performance of different label models. 

\textbf{Future work} $\;$ Future work should give a theoretically precise answer to the reason for why performance deterioration is observed, conduct more extensive experiments to validate that this holds for a variety of datasets and labeling function sets, as well as better characterize the settings in which structure learning actually helps with downstream performance.

\newpage
\bibliography{iclr2021_conference}

\begin{thebibliography}{18}
\providecommand{\natexlab}[1]{#1}
\providecommand{\url}[1]{\texttt{#1}}
\expandafter\ifx\csname urlstyle\endcsname\relax
  \providecommand{\doi}[1]{doi: #1}\else
  \providecommand{\doi}{doi: \begingroup \urlstyle{rm}\Url}\fi

\bibitem[Anandkumar et~al.(2014)Anandkumar, Ge, Hsu, Kakade, and
  Telgarsky]{anandkumar2014tensor}
Animashree Anandkumar, Rong Ge, Daniel Hsu, Sham~M Kakade, and Matus Telgarsky.
\newblock Tensor decompositions for learning latent variable models.
\newblock \emph{Journal of Machine Learning Research}, 15:\penalty0 2773--2832,
  2014.

\bibitem[Bach et~al.(2017)Bach, He, Ratner, and R\'{e}]{structureLearning1}
Stephen~H. Bach, Bryan He, Alexander Ratner, and Christopher R\'{e}.
\newblock Learning the structure of generative models without labeled data.
\newblock In \emph{Proceedings of the 34th International Conference on Machine
  Learning - Volume 70}, ICML'17, pp.\  273–282, 2017.

\bibitem[Bach et~al.(2019)Bach, Rodriguez, Liu, Luo, Shao, Xia, Sen, Ratner,
  Hancock, Alborzi, Kuchhal, R\'{e}, and Malkin]{Drybell}
Stephen~H. Bach, Daniel Rodriguez, Yintao Liu, Chong Luo, Haidong Shao,
  Cassandra Xia, Souvik Sen, Alex Ratner, Braden Hancock, Houman Alborzi, Rahul
  Kuchhal, Chris R\'{e}, and Rob Malkin.
\newblock Snorkel drybell: A case study in deploying weak supervision at
  industrial scale.
\newblock In \emph{Proceedings of the 2019 International Conference on
  Management of Data}, SIGMOD ’19, pp.\  362–375, New York, NY, USA, 2019.
  Association for Computing Machinery.
\newblock ISBN 9781450356435.
\newblock \doi{10.1145/3299869.3314036}.

\bibitem[Boecking et~al.(2021)Boecking, Neiswanger, Xing, and Dubrawski]{IWS}
Benedikt Boecking, Willie Neiswanger, Eric Xing, and Artur Dubrawski.
\newblock Interactive weak supervision: Learning useful heuristics for data
  labeling.
\newblock \emph{International Conference on Learning Representations (ICLR)},
  2021.

\bibitem[Dawid \& Skene(1979)Dawid and Skene]{SkeneModel}
A.~P. Dawid and A.~M. Skene.
\newblock Maximum likelihood estimation of observer error-rates using the em
  algorithm.
\newblock \emph{Journal of the Royal Statistical Society. Series C (Applied
  Statistics)}, 28\penalty0 (1):\penalty0 20--28, 1979.
\newblock ISSN 00359254, 14679876.

\bibitem[De-Arteaga et~al.(2019)De-Arteaga, Romanov, Wallach, Chayes, Borgs,
  Chouldechova, Geyik, Kenthapadi, and Kalai]{biasBios}
Maria De-Arteaga, Alexey Romanov, Hanna Wallach, Jennifer Chayes, Christian
  Borgs, Alexandra Chouldechova, Sahin Geyik, Krishnaram Kenthapadi, and
  Adam~Tauman Kalai.
\newblock Bias in bios: A case study of semantic representation bias in a
  high-stakes setting.
\newblock In \emph{Proceedings of the Conference on Fairness, Accountability,
  and Transparency}, pp.\  120--128, 2019.

\bibitem[Fu et~al.(2020)Fu, Chen, Sala, Hooper, Fatahalian, and
  R{\'e}]{triplets}
Daniel~Y. Fu, Mayee~F. Chen, Frederic Sala, Sarah Hooper, Kayvon Fatahalian,
  and Christopher R{\'e}.
\newblock Fast and three-rious: Speeding up weak supervision with triplet
  methods.
\newblock \emph{ICML}, 2020.

\bibitem[Honorio(2011)]{PGM_Lipschitz}
Jean Honorio.
\newblock Lipschitz parametrization of probabilistic graphical models.
\newblock In \emph{Proceedings of the Twenty-Seventh Conference on Uncertainty
  in Artificial Intelligence}, UAI'11, pp.\  347–354, Arlington, Virginia,
  USA, 2011. AUAI Press.
\newblock ISBN 9780974903972.

\bibitem[Jog \& Loh(2015)Jog and Loh]{misspecified_Gaussian_PGM}
Varun Jog and Po-Ling Loh.
\newblock On model misspecification and kl separation for gaussian graphical
  models.
\newblock In \emph{2015 IEEE International Symposium on Information Theory
  (ISIT)}, pp.\  1174--1178. IEEE, 2015.

\bibitem[Loh \& Wainwright(2012)Loh and Wainwright]{structureLearning2Theory}
Po-ling Loh and Martin~J Wainwright.
\newblock Structure estimation for discrete graphical models: Generalized
  covariance matrices and their inverses.
\newblock In \emph{Advances in Neural Information Processing Systems 25}, pp.\
  2087--2095. Curran Associates, Inc., 2012.

\bibitem[Maas et~al.(2011)Maas, Daly, Pham, Huang, Ng, and Potts]{IMDB}
Andrew~L. Maas, Raymond~E. Daly, Peter~T. Pham, Dan Huang, Andrew~Y. Ng, and
  Christopher Potts.
\newblock Learning word vectors for sentiment analysis.
\newblock In \emph{Proceedings of the 49th Annual Meeting of the Association
  for Computational Linguistics: Human Language Technologies}, pp.\  142--150,
  Portland, Oregon, USA, June 2011. Association for Computational Linguistics.

\bibitem[Ratner et~al.(2016)Ratner, De~Sa, Wu, Selsam, and Ré]{DP}
Alexander Ratner, Christopher De~Sa, Sen Wu, Daniel Selsam, and Christopher
  Ré.
\newblock Data programming: Creating large training sets, quickly.
\newblock \emph{Advances in neural information processing systems}, 29, 05
  2016.

\bibitem[Ratner et~al.(2019{\natexlab{a}})Ratner, Bach, Ehrenberg, Fries, Wu,
  and Ré]{Snorkel}
Alexander Ratner, Stephen Bach, Henry Ehrenberg, Jason Fries, Sen Wu, and
  Christopher Ré.
\newblock Snorkel: rapid training data creation with weak supervision.
\newblock \emph{The VLDB Journal}, 29, 07 2019{\natexlab{a}}.
\newblock \doi{10.1007/s00778-019-00552-1}.

\bibitem[Ratner et~al.(2019{\natexlab{b}})Ratner, Hancock, Dunnmon, Sala,
  Pandey, and Ré]{Multitask}
Alexander Ratner, Braden Hancock, Jared Dunnmon, Frederic Sala, Shreyash
  Pandey, and Christopher Ré.
\newblock Training complex models with multi-task weak supervision.
\newblock \emph{Proceedings of the AAAI Conference on Artificial Intelligence},
  33:\penalty0 4763--4771, 07 2019{\natexlab{b}}.
\newblock \doi{10.1609/aaai.v33i01.33014763}.

\bibitem[Varma \& R{\'e}(2018)Varma and R{\'e}]{Automatic-Lf-gen}
Paroma Varma and Christopher R{\'e}.
\newblock Snuba: Automating weak supervision to label training data.
\newblock In \emph{Proceedings of the VLDB Endowment. International Conference
  on Very Large Data Bases}, volume~12, pp.\  223. NIH Public Access, 2018.

\bibitem[Varma et~al.(2017)Varma, He, Bajaj, Khandwala, Banerjee, Rubin, and
  R{\'e}]{SL-static-analysis}
Paroma Varma, Bryan~D He, Payal Bajaj, Nishith Khandwala, Imon Banerjee, Daniel
  Rubin, and Christopher R{\'e}.
\newblock Inferring generative model structure with static analysis.
\newblock In \emph{Advances in neural information processing systems}, pp.\
  240--250, 2017.

\bibitem[Varma et~al.(2019)Varma, Sala, He, Ratner, and
  R{\'e}]{structureLearning2}
Paroma. Varma, Frederic Sala, Ann He, Alexander Ratner, and Christophe R{\'e}.
\newblock Learning dependency structures for weak supervision models.
\newblock \emph{ICML}, 2019.

\bibitem[White(1982)]{misspecified_MLE}
Halbert White.
\newblock Maximum likelihood estimation of misspecified models.
\newblock \emph{Econometrica: Journal of the Econometric Society}, pp.\  1--25,
  1982.

\end{thebibliography}
\bibliographystyle{iclr2021_conference}

\appendix
\section{Proofs of the theoretical analysis}
\label{sec:proofs}
\vspace{-1.5mm}
\subsection{Problem setup recap}
\vspace{-2mm}
Let $(x,y) \sim \mathcal{D}$ be the true data generating distribution and for simplicity assume that $y \in \mathcal{Y} =  \{-1, 1\}$. Users provide $m$ labeling functions (LFs) $\lambda = \lambda(x) \in \{-1, 0, 1 \}^m$, where $0$ means that the LF abstained from labeling.
We compare two label models, $p_{\theta}$ for the conditional independent case, and $p_\mu$ which models higher-order dependencies: 
\vspace{-1mm}
\begin{align}
    p_\mu(\lambda, y) 
    = \frac{1}{Z_\mu} \exp \left(\mu^T \phi(\lambda, y) \right)
    &=  Z_{{\mu}}^{-1} \exp \left(\mu_1^T \phi_1(\lambda, y) + \mu_2^T \phi_2(\lambda, y)  \right) 
    , \qquad \mu \in \R^{m + M} \\
    p_{\theta}(\lambda, y) &= Z_{{\theta}}^{-1} \exp \left(\theta^T \phi_1(\lambda, y) \right), \qquad \theta \in \R^{m},
\end{align}
where $\phi_1(\lambda, y) = \lambda y$ are the accuracy factors, $\phi_2(\cdot)$ are arbitrary, higher-order dependencies and $ Z_{{\theta}}^{-1},  Z_{{\mu}}^{-1}$ are normalization constants. We assume w.l.o.g. that factors are bounded $\le1$.
Using an unlabeled dataset $X = \{x_i\}_{i=1}^n$ of n data points $x_i \in \mathcal{X}$ to which we each apply the $m$ user provided labeling functions, we attain the label matrix $\Lambda \in \{-1, 0, 1\}^{n \times m}$. With $\Lambda$ we then train the label model to get a set of $n$ probabilistic labels with which we supervise the downstream model $f_w : \mathcal{X} \rightarrow \mathcal{Y}$.

\paragraph{Lemma 1} (Sigmoid posterior). With  $\sigma(x)=\frac{1}{1+\exp(-x)}$ being the sigmoid function, it holds that
\begin{align}
    p_\mu(y \;|\lambda) 
    &=
    \sigma \left( 2\mu_1^T \phi_1(\lambda, y) + \mu_2^T\left( \phi_2(\lambda, y) - \phi_2(\lambda, -y) \right) \right)  \label{eq:posterior1} \\
    p_\theta(y \;|\lambda) 
    &=
    \sigma \left( 2\theta^T \phi_1(\lambda, y) \right). \label{eq:posterior2}
\end{align}

\paragraph{Proof}
\begin{align*}
     p_\mu(y \;|\lambda) 
     &= \frac{ p_\mu(\lambda, y)}{ p_\mu(\lambda)} 
        = \frac{ p_\mu(\lambda, y)}{\sum_{\tilde y\in \mathcal{Y}} p_\mu(\lambda, \tilde y)} \\
     &= \frac{Z_{{\mu}}^{-1} \exp \left(\mu^T \phi(\lambda, y)\right) }{\sum_{\tilde y\in \mathcal{Y}} Z_{{\mu}}^{-1} \exp \left(\mu^T \phi(\lambda, \tilde y)\right) } \\
     &= 
     \frac{ \exp \left(\mu^T \phi(\lambda, y)\right) }{\sum_{\tilde y\in \mathcal{Y}}\exp \left(\mu^T \phi(\lambda, \tilde y)\right) } \\
     &= 
     \frac{ \exp \left(\mu^T \phi(\lambda, y)\right) }{\exp \left(\mu^T \phi(\lambda, y)\right) + \exp \left(\mu^T \phi(\lambda, -y)\right) } \\
     &= 
     \frac{1}{1 + \exp \left(\mu^T \left( \phi(\lambda, -y) - \phi(\lambda, y) \right)\right)} \\
     &=
     \sigma \left( \mu^T \left( \phi(\lambda, y) - \phi(\lambda, -y) \right) \right) \\
     &=
     \sigma \left( 2\mu_1^T \phi_1(\lambda, y) + \mu_2^T\left( \phi_2(\lambda, y) - \phi_2(\lambda, -y) \right) \right)
     ,
\end{align*}
where in the last step we used the fact that the accuracy factors are odd functions, i.e. $\phi_1(\lambda, -y) = -\lambda y = -\phi_1(\lambda,y)$. Eq. \ref{eq:posterior2} follows by the same argumentation.

\subsection{Proof of the bound on the label model posterior} \label{PosteriorBound}
\vspace{-1.5mm}
\paragraph{\emph{Bound}} Our bound on the probabilistic label difference between the two models above is:
\begin{equation} 
    \left| p_\mu(y \;|\lambda) - p_\theta(y \;|\lambda) \right|
    \le
    \frac{1}{2} ||\mu_1 - \theta||_1 + \frac{1}{4} ||\mu_2||_1
\end{equation}

\vspace{-3mm}
\paragraph{\emph{Proof}}
Using Lemma 1 we have that
\begin{align*}
    \left| p_\mu(y \;|\lambda) - p_\theta(y \;|\lambda) \right|
    &= 
     \left|  
          \sigma \left( 2\mu_1^T \phi_1(\lambda, y) + \mu_2^T\left( \phi_2(\lambda, y) - \phi_2(\lambda, -y)\right)       \right)
        -  
          \sigma \left( 2\theta^T \phi_1(\lambda, y) \right)
     \right|
    \intertext{By the mean value theorem it follows that for some c between the arguments of $\sigma$ above} 
    &=
     \sigma'(c) 
     \left|  \left( 2\mu_1^T \phi_1(\lambda, y) + \mu_2^T\left( \phi_2(\lambda, y) - \phi_2(\lambda, -y)                    \right)\right)
            -  
              2\theta^T \phi_1(\lambda, y)
     \right| \\
    &=
     \sigma'(c) 
      \left|  2\left( \mu_1 - \theta \right)^T \phi_1(\lambda, y) + \mu_2^T\left( \phi_2(\lambda, y) -                     \phi_2(\lambda, -y)   \right)
      \right| 
    \intertext{Using the triangle inequality and the fact that $\max_x \sigma'(x) = \max_x \sigma(x)(1-\sigma(x)) = \frac{1}{4}$, we can now bound this expression as follows}
     &\le
     \frac{1}{2}
         \left| 
            \left( \mu_1 - \theta \right)^T \phi_1(\lambda, y) 
         \right|
         +
     \frac{1}{4}
          \left| 
            \mu_2^T\left( \phi_2(\lambda, y) -                     \phi_2(\lambda, -y)   \right)
          \right| 
    \intertext{finally, since the defined higher-order dependencies are indicator functions $\ne 0$ for only one $y \in \mathcal Y$,  and if $||q||_\infty \le 1$ then $|x^Tq| = \left| \sum_i x_i q_i \right| \le \sum_i |x_i q_i| \le \sum_i |x_i| = ||x||_1$,  this reduces to}
    &\le
    \frac{1}{2} ||\mu_1 - \theta||_1 + \frac{1}{4} ||\mu_2||_1.
\end{align*}

\subsection{Proof of the bound on the KL divergence}
\label{KL-div_bound}
\paragraph{\emph{Bound}} 
\begin{equation}
    KL \left( 
    p_\mu(y \;|\lambda) \;|| \; p_\theta(y \;|\lambda) \right)
    \le 
    2||\mu_1 - \theta||_1 + ||\mu_2||_1
\end{equation}
\paragraph{\emph{Proof}} 
We adapt Theorem 7 of \cite{PGM_Lipschitz} to give a bound on the KL divergence between the two label model posterior's. 
First note that with the same line of argumentation as in \ref{PosteriorBound} we have that 
\begin{align*}
    \left| \log p_\mu(y \;|\lambda) -  \log p_\theta(y \;|\lambda) \right|
    &= 
    (\log \sigma)'(c)
      \left|  2\left( \mu_1 - \theta \right)^T \phi_1(\lambda, y) + \mu_2^T\left( \phi_2(\lambda, y) -                     \phi_2(\lambda, -y)   \right)
      \right| \\
     &\le
    2||\mu_1 - \theta||_1 + ||\mu_2||_1,
\end{align*}
where we use that the derivative of $\log \sigma (\cdot)$ is $(1+\exp(x))^{-1} \in (0, 1)$, and is bounded by 1.
Next 
\begin{align*}
     KL \left(p_\mu(y \;|\lambda) \;|| \; p_\theta(y \;|\lambda) \right)
     &=
     \sum_{\lambda \in L} p_{\mu}(\lambda)
    \sum_{y \in \mathcal{Y}} p_\mu(y|\lambda) \cdot 
    \log \left( \frac{p_\mu(y|\lambda)}{p_\theta(y|\lambda)} \right) \\
    &=
    \sum_{\lambda \in L} p_\mu(\lambda)
    \sum_{y \in \mathcal{Y}} \frac{p_\mu(\lambda, y)}{p_\mu(\lambda)} \cdot
    \log \left( \frac{p_\mu(y|\lambda)}{p_\theta(y|\lambda)} \right) \\
    &=
    \sum_{\lambda \in L} 
    \sum_{y \in \mathcal{Y}} p_\mu(\lambda, y) \cdot \left(
    \log  p_\mu(y|\lambda) - \log p_\theta(y|\lambda) \right) \\
    &\le
    \sum_{\lambda \in L} 
    \sum_{y \in \mathcal{Y}} p_\mu(\lambda, y) \cdot \left|
    \log  p_\mu(y|\lambda) - \log p_\theta(y|\lambda) \right| \\
    &\le
    \left( 2||\mu_1 - \theta||_1 + ||\mu_2||_1 \right)
    \sum_{\lambda \in L} 
    \sum_{y \in \mathcal{Y}} p_\mu(\lambda, y) \\
    &= 2||\mu_1 - \theta||_1 + ||\mu_2||_1
\end{align*}

\subsection{Proof of the generalization risk bound}
We now adapt Theorem 1 from \cite{Multitask} to the setting with model misspecification and assume like in \cite{DP, Multitask} that 
\begin{enumerate}
    \item
    there exists an optimal parameter of either $p_\mu$ or $p_\theta$ such that sampling $(\lambda, y)$ from this optimal label model is equivalent to $(\lambda, y) \sim \mathcal{D}$.
    \item
    the label $y$ is independent of the features used for training $f_w$ given the labeling function outputs $\lambda$, i.e. the LF labels provide sufficient signal to identify the label.
\end{enumerate}
For 1. we now assume without loss of generality, that $p_{\mu^*}(\lambda, y ) = p_\mathcal{D}(\lambda, y)$ for an optimal parameter $\mu^* \in \R^{m+M}$, and that we incorrectly use the misspecified label model $p_\theta$.
For the reversed roles (i.e. $p_\theta$ is the true model and $p_\mu$ is misspecified), the following arguments are symmetric. \\ 
Suppose that the downstream model $f_w : \mathcal{X} \rightarrow \mathcal{Y}$ is parameterized by $w$, and
that to learn $w$ we minimize a, w.l.o.g., bounded loss function $L(f_w(x), y) \in [0, 1]$.
If we had access to the true labels, we would normally try to find the $w$ that minimizes the risk: 
\begin{equation}
    w^* = \arg \min_w R(w) =  \arg \min_w  \E_{(x,y) \sim D} \left[ L ( f_w(x), y)\right].
\end{equation}
Since this is not the case, we instead minimize the noise-aware loss:
\begin{equation}
    \tilde w
    = \arg \min_w R_\theta(w)
    = \arg \min_w \E_{(x,y)\sim \mathcal{D}} \left[ \E_{\tilde y \sim p_\theta (\cdot | \lambda)} 
    \left[ L(f_w(x), \tilde y)\right] \right].
\end{equation}
In practice we will get the parameter $\hat w$ that minimizes the empirical noise-aware loss over the unlabeled dataset $X = \{x_1, ..., x_n\}$:
\begin{equation}
    \hat w
    = \arg \min_w \hat R_\theta(w)
    = \arg \min_w \frac{1}{n} \sum_{i=1}^n \E_{\tilde y \sim p_\theta (\cdot | \lambda(x_i))} 
    \left[ L(f_w(x_i), \tilde y)\right].
\end{equation}
Since the empirical risk minimization error is not specific to our setting and can be done with standard methods, we simply assume that the error $| R_\theta(\tilde w) - R_\theta(\hat w)| \le \gamma(n) $, where $\gamma(n)$ is a decreasing function of the unlabeled dataset size $n$.

\paragraph{\emph{Bound}}
By adapting the proof of theorem 1 in~\cite{Multitask} to our case with model misspecification involved, we can bound the generalization risk as follows
\begin{equation}
    R(\hat w) - R(w^*) 
    \le 
    \gamma (n)
    + 2||\mu_1^* - \theta||_1 +  ||\mu_2^*||_1.
\end{equation}
Note that the bound from \cite{Multitask} is mistakenly too tight by a factor of 2 (due to the last step in the proof below that is partly overseen).

\paragraph{\emph{Proof}}
First, using the law of total expectation, followed by our assumptions 2. and 1., in that order, we have that:
\begin{align*}
    R(w) 
    &= \E_{(x', y') \sim D } 
        \left[ R(w) \right] \\
    &= \E_{(x', y') \sim D } 
        \left[  \E_{(x,y) \sim D} \left[ L ( f_w(x'), y) | x = x' \right] \right] \\
    &= \E_{(x', y') \sim D } 
        \left[  \E_{(x,y) \sim D} \left[ L ( f_w(x'),  y) | \lambda(x) = \lambda(x') \right] \right] \\
    &= \E_{(x', y') \sim D } 
        \left[  \E_{\tilde y \sim p_{\mu^*}(\cdot|\lambda)} \left[ L ( f_w(x'),  \tilde y) \right] \right] \\
    &= R_{\mu^*}(w)
\end{align*}
Using the result above that $R = R_{\mu^*}$ and adding and subtracting terms we have:
\begin{align*}
    R(\hat w) - R(w^*) 
    &= R_{\mu^*}(\hat w) - R_{\mu^*}(w^*) \\
    &= R_{\mu^*}(\hat w) + R_\theta(\hat w) - R_\theta(\hat w) + R_\theta(\tilde w) - R_\theta(\tilde w) - R_{\mu^*}(w^*)
    \intertext{since $\tilde w$ minimizes the noise-aware risk, i.e. $R_\theta(\tilde w) \le R_\theta(w^*)$, we have that:}
    &\le R_{\mu^*}(\hat w) + R_\theta(\hat w) - R_\theta(\hat w) + R_\theta(w^*) - R_\theta(\tilde w) - R_{\mu^*}(w^*) \\
    &\le  | R_\theta(\hat w)  - R_\theta(\tilde w)  |
        + | R_{\mu^*}(\hat w) - R_\theta(\hat w) |
        + | R_\theta(w^*) - R_{\mu^*}(w^*) | \\
    &\le \gamma(n) + 2 \max_{w'} | R_{\mu^*}(w') - R_\theta(w') |
\end{align*}
The main term $| R_{\mu^*}(w') - R_\theta(w') |$ we now have in the bound, is the difference between the true/expected noise-aware losses given the true label model parameter $\mu^*$ and the estimated parameter $\theta$ for the misspecified model. We now bound this quantity:
\begin{align*}
    \left| R_{\mu^*}(w') - R_\theta(w') \right|
    &=
    \left|
        \E_{(x,y)\sim \mathcal{D}}
        \left[ 
            \E_{\tilde y \sim p_{\mu^*} (\cdot | \lambda)} 
                \left[ L(f_w(x), \tilde y)\right]
            -
            \E_{\tilde y \sim p_\theta (\cdot | \lambda)} 
                \left[ L(f_w(x), \tilde y)\right] 
        \right]
    \right| \\
    &=
    \left|
        \E_{(x,y)\sim \mathcal{D}}
        \left[ 
            \sum_{y' \in \mathcal{Y}}
            L(f_w(x), y')
            \left(
            p_{\mu^*} (y' \;| \lambda)
            -
            p_\theta (y' \;| \lambda)
            \right)
        \right]
    \right| \\
    &\le
    \sum_{y' \in \mathcal{Y}}
    \E_{(x,y)\sim \mathcal{D}}
        \left[ 
            \left|
            L(f_w(x), y')
            \left(
                p_{\mu^*} (y' \;| \lambda)
                -
                p_\theta (y' \;| \lambda)
            \right)
            \right|
        \right] \\
    &\le
    \sum_{y' \in \mathcal{Y}}
    \E_{(x,y)\sim \mathcal{D}}
        \left[ 
            \left|
            p_{\mu^*} (y' \;| \lambda)
            -
            p_\theta (y' \;| \lambda)
            \right|
        \right] \\
    &\le
    |\mathcal{Y}|
    \max_{y'}
    \E_{(x,y)\sim \mathcal{D}}
        \left[ 
            \left|
            p_{\mu^*} (y' \;| \lambda)
            -
            p_\theta (y' \;| \lambda)
            \right|
        \right] 
\intertext{We can now use our bound from (\ref{BoundProbs}) and get that:}
    &\le
    2
    \left(
    \frac{1}{2} ||\mu_1^* - \theta||_1 + \frac{1}{4} ||\mu_2^*||_1
    \right)
    =
    ||\mu_1^* - \theta||_1 + \frac{1}{2} ||\mu_2^*||_1
\end{align*}
Plugging this back into the term for the generalization risk gives the desired result:
\begin{align*}
    R(\hat w) - R(w^*) 
    &\le 
    \gamma(n) + 2 \max_{w'} | R_{\mu^*}(w') - R_\theta(w') | \\
    &\le
    \gamma(n) +  2||\mu_1^* - \theta||_1 + ||\mu_2^*||_1.
\end{align*}

\section{Factor Definitions}
\label{sec:factor_defs}
We supplement the factor definitions of higher-order dependencies used in this paper. The first two stem from \cite{DP}, the rest we defined ourselves for the conducted experiments, and where motivated by frequently occurring dependency patterns, as the ones in Table \ref{imdb_example_deps}, that are not covered by \cite{DP}. Whenever a factor $\phi_{j,k}(\lambda, y)$ is not symmetric (all factors, besides \emph{bolstering}), we define it so that LF$_k$ acts on (e.g. \emph{negates}) LF$_j$.\\
For the \emph{fixing} dependency we have:
\begin{equation*}
\phi_{j,k}^{Fix}(\lambda, y)
    = 
    \begin{cases}
    +1 &\text{if } \lambda_j =-y \land  \lambda_k = y\\
    -1 &\text{if }  \lambda_j = 0 \land  \lambda_k \ne 0 \\
    0 &\text{otherwise}
    \end{cases}
\end{equation*}
for the \emph{reinforcing} one:
\begin{equation*}
\phi_{j,k}^{Rei}(\lambda, y)
    = 
    \begin{cases}
    +1 &\text{if } \lambda_j = \lambda_k = y\\
    -1 &\text{if }  \lambda_j = 0 \land  \lambda_k \ne 0 \\
    0 &\text{otherwise}
    \end{cases}
\end{equation*}
for the \emph{priority} factor:
\begin{equation*}
\phi_{j,k}^{Pri}(\lambda, y)
    = 
    \begin{cases}
    +1 &\text{if } \lambda_j =-y \land  \lambda_k = y\\
    -1 &\text{if }   \lambda_j =y \land  \lambda_k = -y \\
    0 &\text{otherwise}
    \end{cases}
\end{equation*}

for the \emph{bolstering}:
\begin{equation*}
\phi_{j,k}^{Bol}(\lambda, y)
    = 
    \begin{cases}
    +1 &\text{if } \lambda_j = \lambda_k = y\\
    -1 &\text{if }   \lambda_j = \lambda_k \ne y \lor \lambda_j = -\lambda_k \ne 0 \\
    0 &\text{otherwise}
    \end{cases}
\end{equation*}
and, finally, for the \emph{negated} factor:
\begin{equation*}
\phi_{j,k}^{Neg}(\lambda, y)
    = 
    \begin{cases}
    +1 &\text{if } \lambda_j =-y \land  \lambda_k = y\\
    -1 &\text{if }  (\lambda_j = y \land  \lambda_k = -y) \lor \lambda_j = \lambda_k \ne 0 \\
    0 &\text{otherwise}
    \end{cases}
\end{equation*}

\end{document}